# A Review of Visual Descriptors and Classification Techniques Used in Leaf Species Identification

K. K. Thyagharajan
Department of ECE, RMD Engineering College, Tamil Nadu, INDIA
Email: kkthyagharajan@yahoo.com

I. Kiruba Raji
Department of CSE, RMD Engineering College, Tamil Nadu, INDIA
Email: kiruba161107@gmail.com

## Abstract

Plants are fundamentally important to life. Key research areas in plant science include plant species identification, weed classification using hyper spectral images, monitoring plant health and tracing leaf growth, and the semantic interpretation of leaf information. Botanists easily identify plant species by discriminating between the shape of the leaf, tip, base, leaf margin and leaf vein, as well as the texture of the leaf and the arrangement of leaflets of compound leaves. Because of the increasing demand for experts and calls for biodiversity, there is a need for intelligent systems that recognize and characterize leaves so as to scrutinize a particular species, the diseases that affect them, the pattern of leaf growth, and so on. We review several image processing methods in the feature extraction of leaves, given that feature extraction is a crucial technique in computer vision. As computers cannot comprehend images, they are required to be converted into features by individually analysing image shapes, colours, textures and moments. Images that look the same may deviate in terms of geometric and photometric variations. In our study, we also discuss certain machine learning classifiers for an analysis of different species of leaves.

## 1 Introduction

Plant classification is an active research area, with plants being used in agriculture, medicine and food industry, as well as in the preparation of cosmetics and a range of food products. Individuals cannot characterize plants as effectively as botanists, who do so by classifying those utilizing leaves, flowers, seeds, and roots. Today, however, all vegetation needs to be digitized, owing to the ecological conditions prevailing. Agent-based systems classify plants into species that can be used in medicine and as food. Keeping in mind the end goal, which is to provide data on therapeutic plants, it is critical to have an intelligent system framework that recognizes natural species with the assistance of their digitized databases.

An intelligent system is a key strategy utilized in plant-based recognition systems to create real models from plants, incorporating pattern classification and object recognition. Researchers have created a plant acknowledgment framework utilizing plant leaves, flowers, fruits and seeds and by taking into consideration the visual content of their images such as color, texture, and shape. Nevertheless, such a framework does not help users who need to discover particular image objects.

Consequently, researchers must use object detection and object recognition techniques for a domain-specific object search. Domain-specific image searches can be classified into narrow and wide [1]. Narrow image domains, more often than not, offer restricted variability and better comprehension of the visual substance of images. Wide image spaces, on the other hand, have high variability and consistency for basic semantic ideas of images.

Object recognition, a process of identifying objects based on their appearance and features, is applied to domain-specific object searches. Appearance-based object detection uses images and a range of conditions like changes in size, shape, color, lighting and viewing direction. Extracting effective features is fundamental to identifying objects in appearance-based object detection.

This paper is organized as follows. Section 2 deals with the overview of species identification System. Sections 3, 4 and 5 deals with the various visual feature extraction techniques needed to recognize leaf images based on leaf shape, texture and scale/rotation invariant techniques.

Section 6 examines the classification techniques for classifying leaves. Section 7 discusses a combination of the different features and classifiers for effective classification in the different leaf databases.

## 2 An Overview of Species Identification Systems

The researchers Du et al. [2] have analysed morphological features and invariant moment features of various shapes of different plant databases and applied the move median centres (MMC) hyper sphere classifier to classify leaf species. They used a leaf database containing only a single leaf image against a blurred background, and collected a total of 20 species of different images with a total of 400 scanned leaf images. Macleod et al. [3] investigated several computer-assisted systems for the species identification of living and non-living things based on the DNA bar-coding scheme. They studied systems in oceanographic-based research and pale ontology, and tested his work in the Digital Automated Identification System (DAISY), classifying only 30 species. They worked on dinoflagellate categorization using the Artificial Neural Network (DiCANN) system to identify phytoplankton species with 72% accuracy. Pattern Analysis, Statistical Modelling and Computational Learning (PASCAL) were used to classify common objects.

A plant species identification system for the broad leaves found in Norway was proposed by Babatunde et al. [4] which were based on the morphological features of the leaves and they also discussed different features of leaves and feature extraction techniques. In [5], various leaf structures and flower feature extraction techniques and problems in an agricultural environment were reviewed. Detailed information of the important survey papers with their references and number of citations based on Google Scholar as of June 2017 is presented in the Table 1.

We have selected 200 papers with different leaf databases for our study, based on the following paradigms (P1– P7). Our search identified papers that discussed only feature extraction techniques, as well as those that included classification techniques, those based on particular leaf species, those that included a combination of shape and venation, those that included a combination of texture and texton, and those that worked to resolve the problem of big data. We list here the paradigms used for our study, and Table 2 shows the number of papers included for this detailed analysis so as to handle different problems in agricultural research.

P1: Analysis based on different leaf shapes

P2: Analysis based on venation

P3: Analysis based on leaf tip, base, and margin

P4: Analysis based on texture/ texton

P5: Analysis based on moment invariant descriptors

P6: Analysis based on different classification techniques to resolve problems with inter and intra-class classification, imbalanced data, and managing big data.

P7: Analysis based on different leaf databases

### 2.1 Block Diagram of Leaf Recognition System

The general block diagram of leaf species identification system is shown in Fig. 1. In this system a user gets the leaf image to be identified. Then the system performs image pre-processing such as conversion of a colour image to grayscale image, image smoothing by removing noise, segment the images etc. Next, the system extracts the general features of leaf such as shape, colour, texture and some of the leaf specific features such as leaf tip, base, apex and margin and venation information. These features are compared with the features of the leaves stored in a database to identify the species of the leaf based on Intra and inter classes' similarity. Table 3 shows some of the leaf recognition systems published.

## 3 State-of-the-Art Techniques in Feature Extraction

A feature is a piece of information relevant to a specific leaf image, and is divided into two types: local and global. Local features are extracted from leaf patches and global features from leaf shape, texture and colour. All leaves are identical in terms of colour, which can vary with climatic changes. Colour, shape and texture are appropriate features for the classification of leaf species. There are two types of leaves: simple and compound leaves, according to leaf manual [10] their general structures are as shown in Fig. 2. Cope et al. [5] discussed the morphological structure of simple leaves, which are identified through key features, such as color, shape, margin, venation, and arrangement. Compound leaves, however, are identified by the number of leaflets in a stalk, with the extraction of feature from single leaflet. There is, therefore, a need for appropriate features for the identification of leaf species. Sharma and Gupta [11] presented an overview of some of the common methods used for leaf feature extraction and classification.

### 3.1 Feature Extraction Techniques

Feature extraction is an important technique used in image classification, pattern recognition and object recognition. In order to have effective classification of plant species researchers should decide to extract efficient features. Researchers classify plants using roots, fruits, seeds and flowers [12–14]. Leaf colour [15] cannot be considered a viable feature for classification because it may vary with climatic and camera calibrations. Given that most leaves are green, they are to be classified through shape, texture and invariant feature descriptors that are invariant to translations, rotations and scaling transformations of images. Since colour is not considered, grayscale images of leaves are used for identification. Figure 3 shows different feature extraction techniques.

Table 1 Citation details of review papers

| Authors | Journal | Topic discussed | Years taken | No. of references | No. of citations |
|---|---|---|---|---|---|
| Du et al. [1] | Applied Mathematics and Computation | Leaf shape analysis using morphological features of leaves | 1993–2004 | 20 | 284 |
| MacLeod et al. [3] | Article in Nature | Automatic species identification in both living and non-living things | 1959–2010 | 10 | 130 |
| Cope et al. [5] | Expert Systems with Applications | Review of digital morphmetric analysis of leaf shape, texture, and flower analysis | 1992–2012 | 113 | 154 |
| Babatunde [4] | Journal of Agricultural Informatics | An outline of a computer-assisted system for plant species identification | 2003–2012 | 27 | 4 |
| AbJabal [187] | Journal of Computer Science | A review of different feature extraction and classification technique | 2003–2011 | 26 | 22 |
| Waldchen et al. [6] | Archives of Computational Methods in Engineering | A review of local, global and moment invariant analysis in different leaf databases | 2006–2016 | 159 | 1 |

Table 2 Inferences of data sources

| Visual descriptors of leaf and classification | No. of papers selected |
|---|---|
| Geometrical descriptors | 30 |
| Leaf shape/tip/base/venation | 40 |
| Texture/texton | 25 |
| Invariant descriptors | 15 |
| Classification and leaf databases | 80 |
| Survey | 7 |
| Leaf identification system | 3 |

Table 3 Leaf recognition system

| Author | Leaf identification | Features and classification technique | System type |
|---|---|---|---|
| Pauwels et al. [7] | Computer assisted tree taxonomy | Leaf shape<br>Nearest neighbor classifier | Web service |
| Pharm et al. [8] | Computer aided plant identification system | Leaf margin<br>HOG ? Hu feature<br>Support vector machine | Computer based system |
| Rajeb Sfar et al. [9] | Plant system based on botanical idkeys | Taxonomy and landmarks act as botanical id key | Computer system |

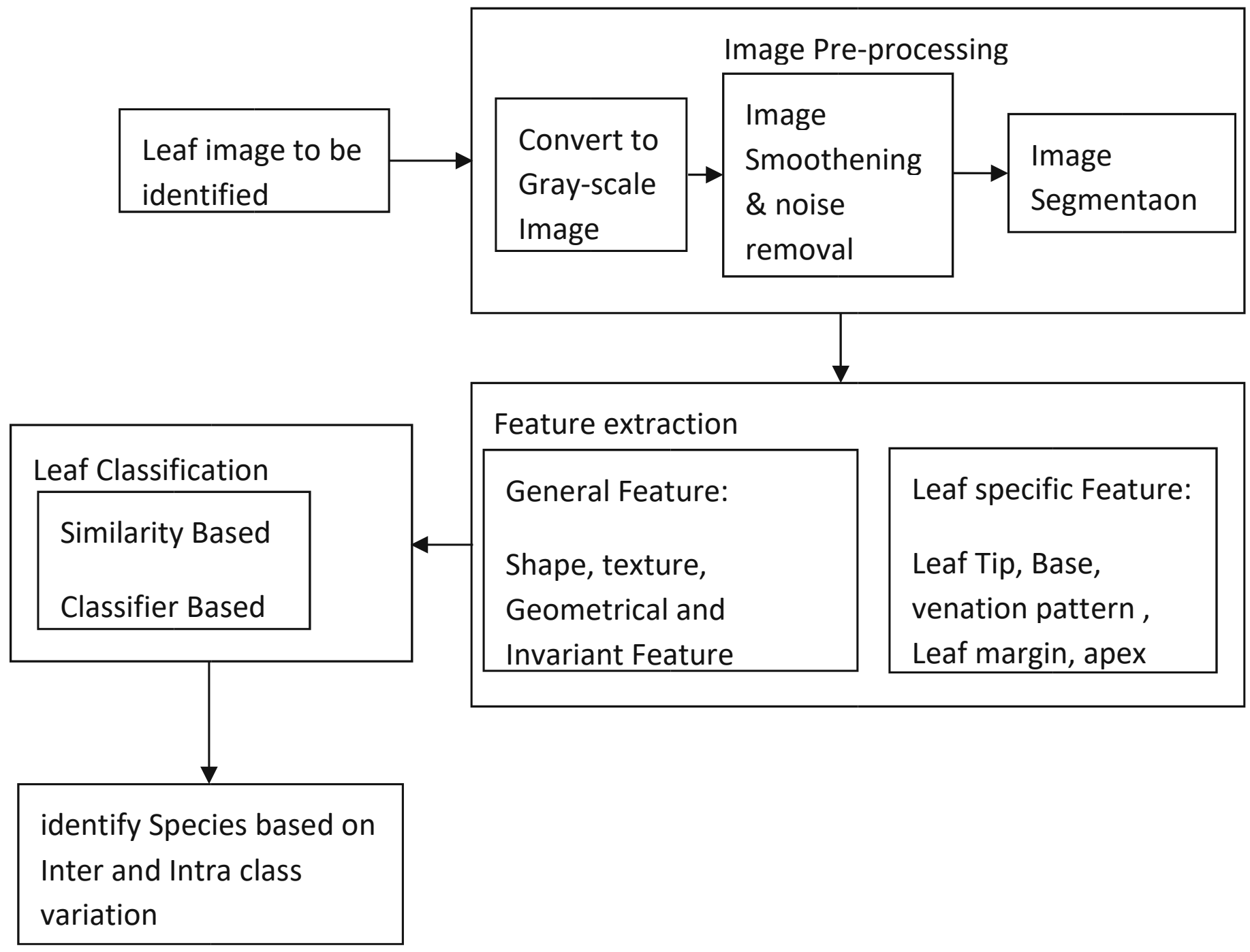

Fig. 1 Block diagram of leaf species identification system

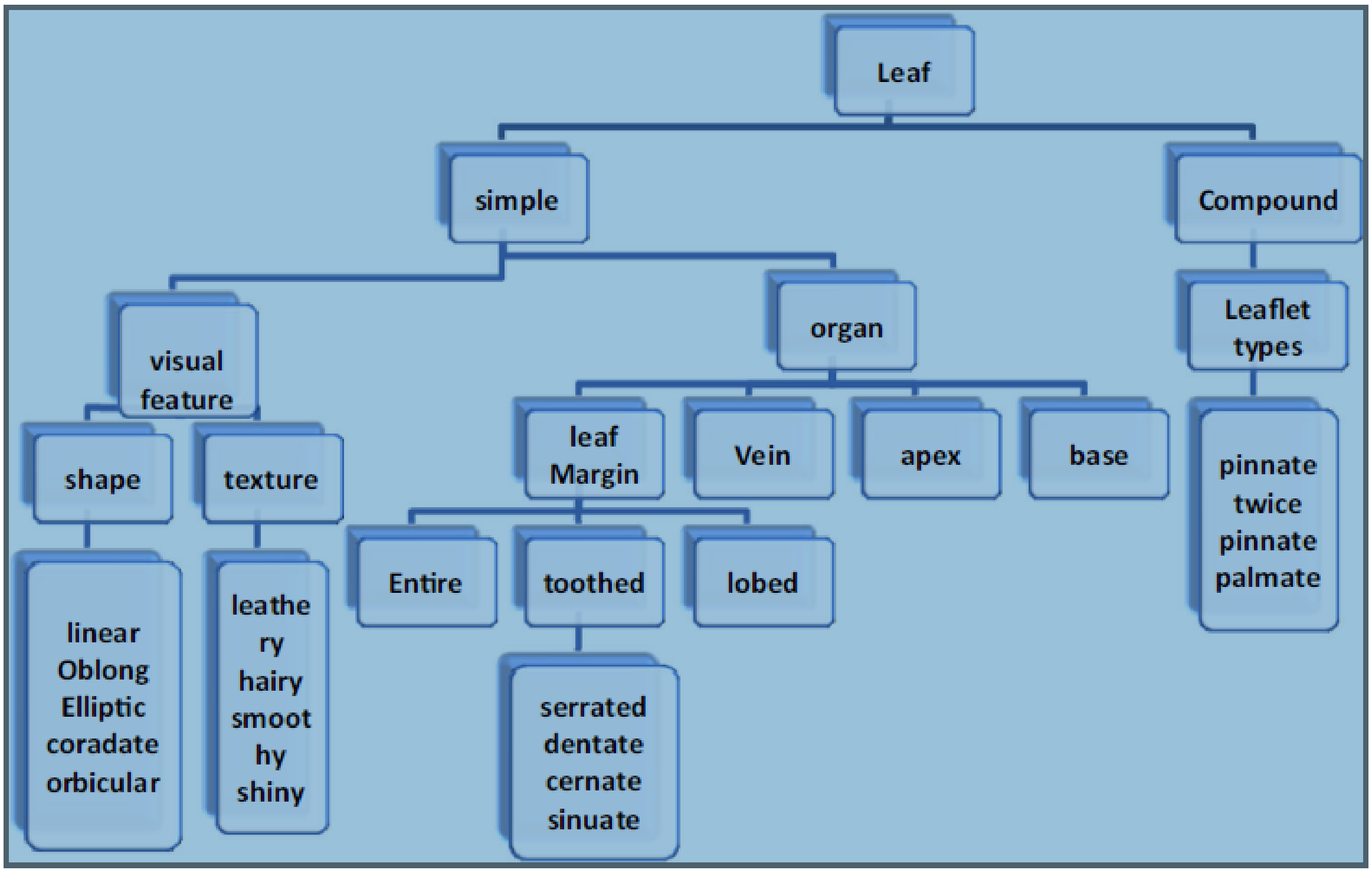

Fig. 2 Types of leaf features

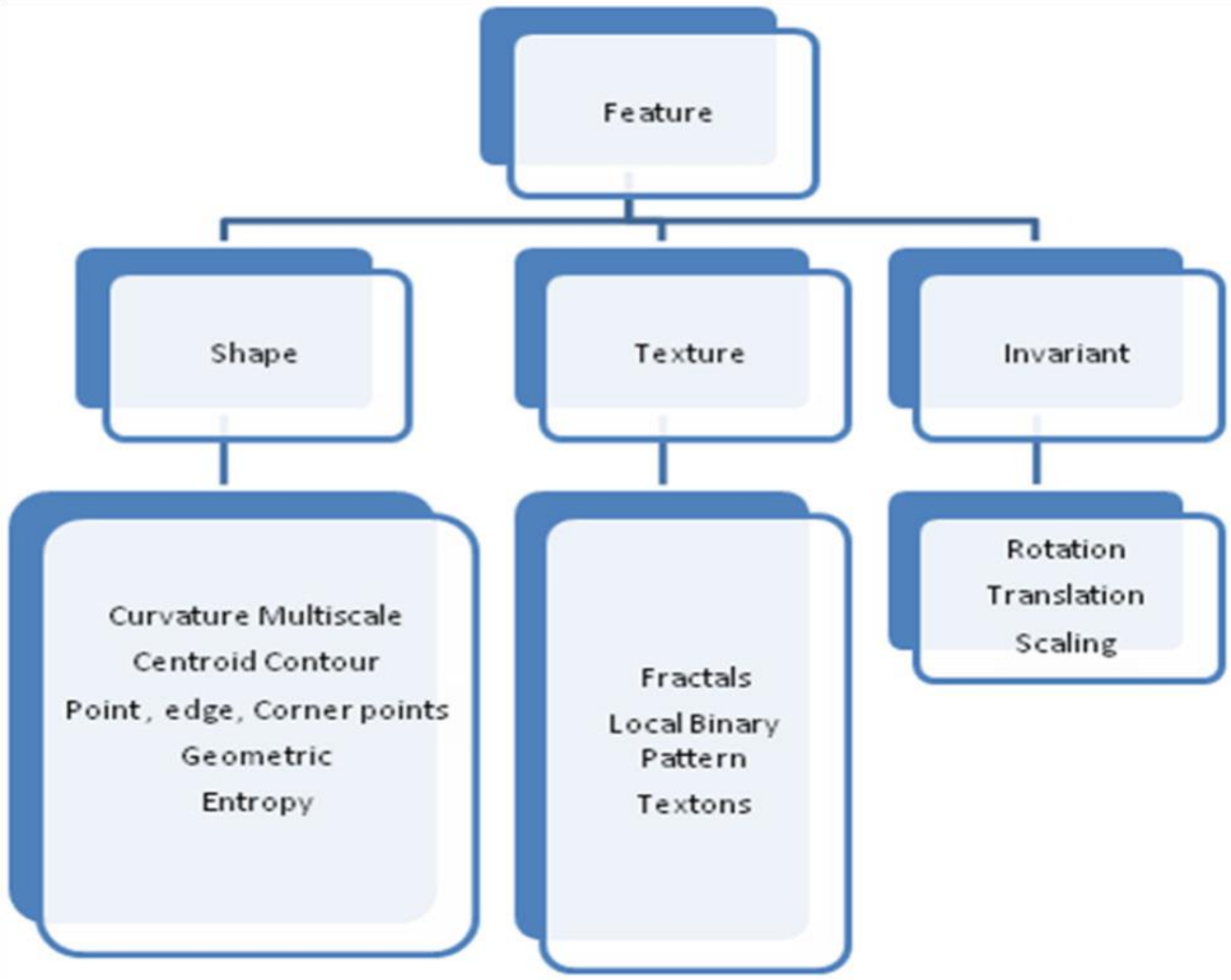

Fig. 3 Feature extraction techniques

### 3.1.1 Curvature Descriptors

Curvature Scale Space (CSS) is a technique used to measure the contours of shapes, extracts the concavity and convexity of curvature. It is invariant to translation and rotation in a viewpoint direction but not in scale, because it varies with the Gaussian kernel (a) and cannot easily fix the value of the Gaussian kernel. It leads to misclassification of serrated and lobe-shaped leaves. Curvature is a vital property of leaves and curvatures are computed using differential techniques. However, it produces more noise, is sensitive to rotation, and generates different feature vectors with different scales. It is impossible to sustain all the curvature features combined together in one feature vector. Aligning them all in one particular point is a difficult task, because the features differ for each scale.

According to [16] CSS used to identify the starting and Fig. 3 Feature extraction techniques ending points of the venation feature points of leaves by estimating the maximum angles of the leaves. The densities of feature points are estimated using the Parzen window method for non-parametric density estimation and it can be applied to any data distribution. We cannot, however, get to choose the correct window size. According to [17], since veins are represented as strings used for semantic interpretation, there is no need to find the starting and ending points. But these methods cannot be used for imperfect and overlapped leaves. Grinblat et al. [18] used an unconstrained hit or miss transform technique to extract particular patterns in foreground and background pixels. When applied to leaf images, central vein patches are extracted from leaves and various geometric features are calculated for the veins. The SIFT descriptor [19, 20] were used to extract key features from an image. It produces good results on the circular orientation of an image, and is well suited to illumination and various viewing conditions. It extracts histogram features from local patches. The authors extracted corner points using Mean Projection transform (MPT) instead of CSS, it produces indistinguishable variations as well as aliasing. To eliminate such problems, the Mean Projection Transform extracts corners that have high curvature. The Flavia dataset produces accuracy of 87.5%.

The researchers Chen et al. [21] proposed a velocity representation technique to represent curvature points. This algorithm computes only 9 points on a leaf contour. It reduces the running time of the algorithm, because the CSS computes 200 intersection points on the curvature and increases the running time. Square root velocity representation [22] was used for shape-based leaf classification to solve the intra-class and inter-class variability of leaf images. It automatically detects similarities by computing the geodisc distance of statistical shape features and 2D planar curves by computing the elastic deformations of the Riemannian structure. It is applied to the Flavia leaf dataset.

### 3.1.2 Multi-scale Descriptors

The multi-scale descriptors furnish much more information about leaf contours. Derived from the scale space and image pyramid structure, it extracts image features at various levels by capturing local and global features from low- to high-resolution scales. It provides the maximum discriminating power and is robust to noise depending on the boundaries of leaves and not the regions of an image. As a result, it works well on feature space rather than image space. Multi-scale Triangular Area Representation (MTAR) is used in [23] which is affine invariant, robust to noise and provides the features of images concavity and convexity. He also developed triangle side length and triangle-oriented angle descriptors for leaf images. The researchers Wang et al. [24] introduced Multi-scale Arc Height Descriptor (MARH) which is invariant to translation. It enumerates a local normalization technique for each scale to employ rotation and scaling, because the local normalization rendered for each scale is based on the maximum value of arch height descriptors. It leads to shape dissimilarity at each different scale, so is invariant to translation and scaling. It measures the arch height of palmate-shaped and lobe-shaped leaves but is unsuitable for overlapped leaves. In this method, the local normalization scheme is applied for scaling and rotation. It takes longer execution time, compared to other invariant descriptors.

A new method called the Multi-scale Bending Energy (MBER) was proposed by Souza et al. [25] which require energy to perform at the lowest energy rate on a curvature signal based on its sensitivity to the local features of the shape contour. It provides low noise immunity and spatial locations of certain prominent points. Given these limitations, its use in shape description is rather limited. Researchers of papers [26] used curvelet transform, which is a multi-scale object representation technique applicable only to objects with small length scales. It is not applicable to natural images—for, while increasing image size, the edges end up looking like straight lines. This property is not suited to natural images of leaves and flowers, and is only applicable to text and cartoons. Multi-scale R-angle [27] descriptor, compared to all the other descriptors, is intrinsic to shape contours under translation, rotation and scaling, because the other methods need normalization for scaling.

### 3.1.3 Centroid Contour and Angle Code Descriptors

The Centroid Contour Descriptor (CCD) used by Sangle et al. [28] measures the distance between the centre and the boundary points, and is invariant to translation and rotation. If a user knows the location of the starting point, the image produces the same shape signature for the rotated images. The Angle Code Descriptor (ACD) computes the continuous orientation angles of leaf shapes but provides limited shape information. So they combined both CCD and ACD to retrieve all the essential information of a leaf image and applied these methods to the mango, tulsi, rose and Asoka tree species. The CCD and ACD were used to extract, oblong and orbicular leaf shapes and to identify leaf species in [29]. Knight [30] developed android app for identifying 6 different classes of leaves. He used CCD and ACH for extracting leaf features. Thangirala [31] proposed CCD with Centroid Contour Gradient for broadleaf classification and used CCG to extract leaf gradients between two points on the leaf's contour. These points were used to measure the angles between the tip and the base. Bong et al. [32] suggested to normalize the tip and base of the leaf and used centroid contour gradient (CCG) to capture the curvature of the tip and base of the leaf. They achieved 99.47% classification accuracy by using feed-forward back propagation network as classifier Fotopoulou et al. [180] advised to convert the centroid contour distance and angle code sequence into 1D time delay sequence and he measured similarity of leaf shapes through Multidimensional Embedding Sequence Similarity (MESS).

### 3.1.4 Point and Edge-based Feature Descriptors

A new descriptor called the shape context was introduced in [33] to dissociate shape information from different shapes. It is a technique used to extract point information from a shape's contours, measure similarity differences between feature vectors of various points in an image, and isolate information from the neighbouring pixels of an image. The transformation of an object does not affect shape context information. It is invariant to rotation since it performs log polar operations while computing shape context information. It is invariant to small affine transformations, occlusions, the presence of outliers, and is applicable to clear images. Shape context is used to calculate the local and spatial information of an image. In [34], an advanced shape context method was introduced to reduce computational cost. In this method they used two sets: a voting set and a computing set. While the voting set was used to build the histogram information of the shape, the computing set was used to compute the shape context information of various shapes. This method was used for polygonal shaped leaf images.

The researchers of paper [35] proposed a new technique in shape context termed the Inner Distance Shape Context (IDSC), where the Euclidean distance is used to compute the cost matrix between two shapes. But it does not consider how many line segments are crossed in shape boundaries and, further, increases the computational cost. The technique solves the problem above by calculating the length of the shortest path with in shape boundary, and is invariant to articulation points that requires complex matching algorithm to compare a set of points.

The inner distance shape context (IDSC) technique was proposed in [36] for articulated shape recognition and it is a very useful technique when the veins in leaves are damaged. The IDSC cannot store information on compound and serrated leaves or model the local details of leaf shapes well. It models only global information and misses some local information. Zhao et al. [37] introduced the Interdependent Inner Distance Shape Context (I-IDSC) to calculate the shape context with different aspects, but different plant species can have a common shape and the I-IDSC discriminates between leaves with similar shapes but different margins. It accurately classifies both simple and compound leaves, retains the most discriminative information, is very fast and offers cheap storage.

A Histogram of Curvature over Scale (HoCS) [38] is method to measure histogram features in one single point because it is simple to compute, compact and requires no alignment. It is a multi-scale invariant integral curvature measure calculated from circle-cantered point. It gives natural notions of scale by resizing the image in segmented areas. It is robust to noise and invariant with rotation. It also removes holes in leaf images, extracts curvatures from boundaries, and measures smooth as well as serrated margins. This technique was used in the paper [39] to extract the arc and area features of lobe-shaped leaf margins, but it is not suitable for all leaves. This technique was also used for Costa Rican species as well in [40]. The HoCS, however, is not articulation invariant. An active shape model was proposed in [41] to find edge points and leaf tip points by overlapping two leaf points and tracing their continuous shape. The model was used for slender and thread-type leaves. An active polygonal model technique was used by Cerutti et al. in [42] to extract the tip and base information of a leaf by computing 10 feature points such as the base, base angle, tip of the angle and the isosceles triangle. This model fits polygons on images, helps to preserve corners, and extracts information on leaf tips and bases. Cerutti et al. [43] represented the contour of the leaf margin as a sequence since the leaf margin is the most discriminated feature of a leaf. Toothed leaf margins are represented as a string. This method presents information on leaves semantically, and is most useful, especially when the leaf is unavailable at a time. The drawback, however, is the danger of misclassification of the leaf margin when the margin in question is imperfect.

Du et al. [44] presented a leaf species identification method using shape matching technique. They adopted Douglas–Peucker approximation algorithm to get the attributes of the leaves and proposed a modified dynamic programming (MDD) algorithm for shape matching. This method is suitable even if the leaves are overlapped, distorted and partial. It works with any number of dimensions and extracts a small number of points by splitting the entire contour into small curves. It depends on the starting point, and is a pure geometrical algorithm to obtain a smaller number of vertices. It also affects from noisy images.

### 3.1.5 Edges and Corner Points

Edges are significant features of leaf images in terms of measuring sharp variations in images. The Sobel edge detection operators were used to extract edge features from images in [45]. From the edges, feature points were found which intersect the edges and achieved 100% accuracy with 13 different plants. The model ascertains damages to veins. Corner features [46] are useful to find the similarity of leaf images because corners are intersections of two different edges or interest points under various different directions and lighting conditions. They are stable across different sequences, useful when there is damage to the corners, and are the same for all leaves. Harris Corner detectors are used to find the different directions of contours directly. The angles are arranged in ascending order, stored in an array and compared to find out the least angle of the unmatched image. Tekkesinoglu et al. [193] used morphological transformation and edge detection techniques to identify the leaf boundary of overlapped (Hevea leaves) rubber tree leaves.

### 3.1.6 Leaf Tooth, Tip, Margin

A tooth is a depth incised towards the sinus and it is different from a lobe. In [47], the authors estimated the tooth's area, perimeter and internal angles for the whole tooth of Tilla trees by applying the tooth-finding algorithm. They found the points on edges by calculating the centroid distance from the center to the edge and thereafter marked the sinus of the margin. Each tooth can be represented as a triangle containing a tip and sinus on both sides. They used LDA to classify the species of Tila family such as Tilla platypyllus, Tilla Americana and Tiila Tomentosa, and achieved a classification accuracy of 68.3%. Susan corner detectors were applied to detect leaf image corners and Non-leaf image corners are removed using Pauta Criteria in [48]. The leaf number, leaf rate, leaf sharpness and leaf obliqueness of leaf tooth features are measured and the leaves are classified using the sparse representation of leaves. The tangential angle approach was used in [10] for finer angular details of the leaf boundary. Nandyal and Govardhan [194] used geometrical distances such as mid vein length, apical extension length, basal extension length and leaf length to measure base angles and apex of different shapes of leaves and they used curvature scale space for measuring margin coarseness.

### 3.1.7 Geometric Features

Geometric features are used for leaf classification because they are of low-dimensional compared to other features, incur low computational cost and take less time to extract the features. Morphological features were used for weed classification by Cho et al. [50]. Singh et al. [51] used 5 basic geometrical features and 12 digital morphometric features with Fourier moments to classify 32 different plants. Wu et al. [52] proposed slimness, roundness, and solidity, and moment invariants to classify plant species. In [53] Dornbusch and Andrieu recommended the Lamina 2 Shape algorithm to analyse lamina-shaped gross leaves to measure their length, width and area. They estimated the accuracy of the width by calculating a predefined lamina shaped model. This algorithm forms equally-spaced perimeters on the area of the leaf and is not suitable for all types of leaves. The Waddle disk diameter method was used to measure the roundness of leaf area for grass-like species such as ryegrass, wheat and brome grass in [54]. Hossain and Amin [55] used biometric-based geometrical features of leaves for broad and flat leaves by selecting reference points from leaf blades and leaf bases.

The researchers Wu et al. [56] proposed 5 geometric features—diameter, physiological length, physiological width, area, and perimeter—and 12 morphological features including smoothness, aspect ratio, form, rectangularity, narrowness, perimeter ratio of the physiological length and width, and 4 vein features of the leaf. These features were used to recognize 32 different kinds of plants. Tzionas et al. [57] proposed morphological features of leaves to classify different species of leaves. Kadir et al. [58] used geometric features such as slimness and roundness, to measure the regularity of leaf shapes, and dispersion to measure their irregularity. These features were tested on the Flavia dataset to classify the leaves. In [59] the authors applied digital morphological features to classify 32 different plant species and rate them. Singh et al. [60] observed that a minimum of 7 morphological features of elliptic-shaped leaves are essential for feature extraction. Geometrical and morphological features were used in [61] to classify compound leaves. Instead of extracting the features of the whole leaf image, the authors successfully extracted geometrical features from each leaflet of an image of clustered potato and tomato leaves. Kaif and Khan [65] used geometrical and shape-defining features such as the shape of the object, sets of horizontal and vertical lines, endpoints, boundary points, slopes between two lines and Fourier descriptors for the TRS invariant features. The authors of paper [66] used the morphological covariance method to extract coarseness, anisotropy, and textural data of images. They used structuring elements to represent the contour of curves, extracted edges from the leaf contour, and extracted shape information from images and introduced the Circular Covariance Histogram to extract venation information from leaf images using the circular structuring element. Statistical features were used to extract deformable objects by Chaki et al. in their paper [67]. They divided the leaves into equal parts and calculated the statistical features separately, as both deformable and whole leaves have the same structure, so features extracted from one place are used as a vector for deformable objects. Dutta et al. [72] used geometrical and morphological characteristics of leaves to classify mango plants. Most researchers [62–64, 68–71] use geometric features for leaf classification, alongside weed detection because of the fewer dimensions involved, but they do not consider details of leaf margins. Leaf margins contain most of the details, and are only applicable to smoothed leaves Manik et al. [200]. used morphological features of Anthocephalus cadamba to identify diseases in leaves.

### 3.1.8 Entropy and Pulse Coupled Neural Networks

The pulse-coupled neural network, an artificial neural network model, is used to extract features from leaf images. The image size and neural network size are same. Pulse Coupled Neural Network was used by Wang et al. in [73] to classify leaves using the entropy sequence with Hu and Zernike moment invariants. Liu et al. [74] used an adaptive unit-linking PCNN to extract the centre of the time matrix from images. Different from the PCNN an Intersecting cortical Model (ICM) used by Wang et al. [75] acts as an anti-agent for noise and anti-geometric properties of images. Table 4 summarizes the feature extraction techniques, the features extracted from leaves using those techniques and, the advantages and disadvantages of those techniques.

## 4 Texture of Leaf

An image texture is recognized by a set of metrics designed to quantify the perceived texture of an image. It gives us information about the spatial arrangement of colour or intensities in an image or a selected region of an image. Image textures, which can be artificially created or found in natural scenes captured in an image, can be used to classify images. A texture-based feature extraction method extracts the characterization of regions in a leaf image by means of its texture content such as smoothness, roughness or silkiness. The texture of leaves differs for the same species of leaf.

### 4.1 Texture Features Based on Fractals

The topological structure is used to measure how close two objects are to each other. In [80], the authors used a Lie group of region structures to measure the texture of weeds and provide information about pixel intensity and spatial features of broadleaf weeds. The smooth manifolds of local symmetries were derived at by applying the Riemannian Manifold on the leaf surface. The dimension of a region covariance of the leaf surface is lower than that of the original image. It extracts multiple features such as information on edges and directions. Fractals measure the self-similar texture of leaves as well as the roughness of the leaf surface. A multi-scale Minkowsi fractal dimension method was used to analyse and recognize leaf images in [81, 82]. This method extracts outline and vein features as curves. Usually, objects and patterns have distinct geometric natures in fractals and, in order to overcome this difficulty, they used the multi-scale Minkowsi fractal dimension technique for classifying Passiflora leaf morphometry. In [83], the researchers used new fractal refinement technique for classifying species based on contour, contour nerves, nervure fractals of three different levels. Mutchar and Fatichah [84] used lacunarity feature for leaf classification as the fractal dimension cannot discriminate between two objects with different patterns/texture. It measures the spatial distribution of gaps with certain image textures. Casanova et al. [85] used Gabor filter to extract texture features from images. It collects various image features and extracts energy signature from leaves. He evaluated 20 different classes of Brazilian flora using Linear Discriminant Analysis and achieved 86.00% of classification accuracy.

Vijayalakshmi et al. [86] extracted texture using Gabor filter with 30-degree rotation angle in a 5 9 5 pixel neighbourhood and obtained 13 different structural characteristics of a leaf compared to other kernel-based methods that use a 45-degree rotation angle to extract only 8 different statistical measures. Boligond–Minkowski fractal dimension method was used in [87, 88] to count the number of boxes in a spatial relationship of pixels for the classification of the Brachiaria species. But it does not obtain any invariant features. Singular value decomposition method was directly applied [89] on a real matrix to classify texture characteristics with high-level factorization and provides good results in varying lighting conditions. A gray-tone spatial dependency matrix [90] and LBP patterns were applied for the classification of medicinal leaves. The Local Gabor phase quantization (LGPQ) scheme proposed in [91] to extract different features of texture changes gradually along with a rich set of discriminated information because of the magnitude of information it carries. The authors extracted the entropy, mean, skewness, standard deviation and variance.

### 4.2 Local Binary Patterns Based on Texture

The Local binary pattern (LBP) is an image feature, which transforms image into an array of values. It describes about the changes in the neighbouring pixels. Qi et al. [92] introduced a pair-wise rotation invariant co-occurrence local binary pattern (PRICoLBP) and applied to colour images. It represents the local curvature as well as edge contour information. This technique was applied to various databases comprising flower and leaves. An LBP histogram Fourier feature (LBP-HF) [93] identifies uniform patterns using Fourier descriptors. It stores all uniform patterns in a single bin and the authors used all the information on pixels, leaf interiors and exteriors separately. A modified local binary pattern was proposed in [94], where LBP binary values are calculated based on thresholding. It lends same LBP code for two different patterns. To overcome this problem the mean and standard deviation of the neighbouring pixels were taken into account. It captures the structural relationship between the gray values of the pixels in the neighbourhood. The LBP was combined with the gray-level co-occurrence matrix in [95] for tea leaf classification. In the basic LBP, every pixel needs to be calculated for obtaining LBP values, and computing the LBP is a time-consuming process. To circumvent the problem, the authors introduced a non-overlap window that includes a centre pixel and its neighbour pixels in a single gray-level image. There is no overlap between the windows in this technique. Since the GLCM is used to calculate the relationship between two windows, it produces multiscale texture features.

A multiscale local binary pattern was applied on the path integral (pi-LBP) in [96]. In all multiscale LBPs, local information is encoded individually in each scale, but the pi-LBP can effectively encode the cross-scale correlation and provides better texture description. A pixel-based LBP was introduced in [97] instead of computing global information built on a block-based LBP the authors computed LBP based on center pixel of a half-size window which determines how much local and global information is included in the texture descriptor. It produces powerful relations for the intra-class variability of textures. Sumathi et al. [98] used Gabor filter for textural, statistical and spatial frequency domain relationships in leaf classification. The LBP variance [40] was applied to classify Costa Rican plant species. It detects micro texture veins as well as areas between veins and reflections. It returns a histogram of features and counts the position in which it corresponds to the particular leaf texture which has an LBP code. The gray-level co-occurrence matrix was used for herb detection in [99].

Table 4 A summarization of leaf shape/tip/base/venation points and edge-based and geometric descriptors

| Feature extraction technique | Image | Extracted feature from leaf | Pros and cons of feature extraction technique |
|---|---|---|---|
| Shape context [33] | 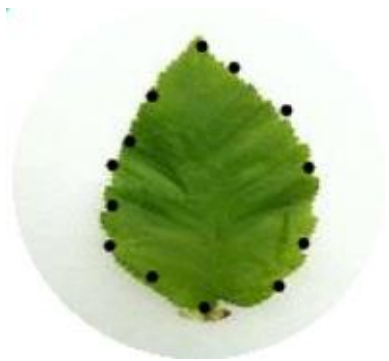 | Point information from shape contours | Isolates information from nearby pixels, and is invariant to affine transformation, occlusions, and the presence of outliers<br>Applicable to only unaffected images |
| Advanced shape context [34] | 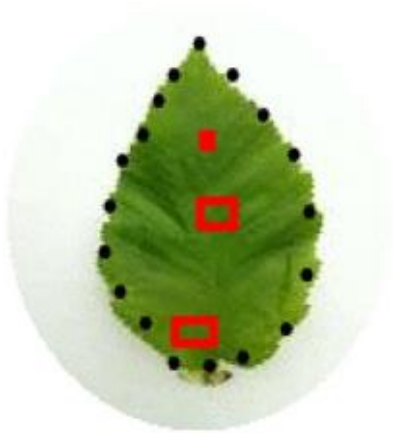 | Relations between salient and margin points | Reduces computational costs<br>Applicable to polygonal objects |
| Shortest-path texture context [35]<br>Inner-distance shape context [36] | 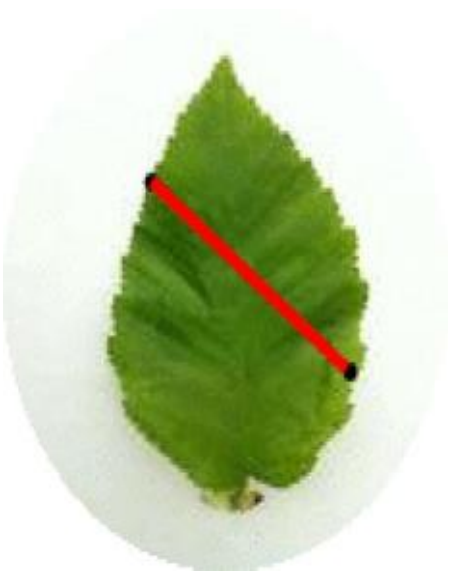 | Leaf vein | Measures the relative orientation along the shortest path<br>Used for texture nonuniform illumination changes of leaf veins<br>Useful when veins are damaged and models only global information<br>Cannot store information on compound and serrated leaves |
| Histogram of curvature over scale (HOCS) [38–40] | 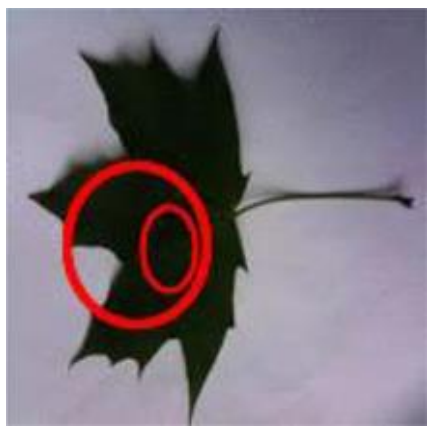 | Histogram information in one single point | Robust to noise and rotation invariant<br>Only suitable for lobe-shaped leaves<br>It is not articulation invariant |
| Douglas Peucker contour Approximation [44] | 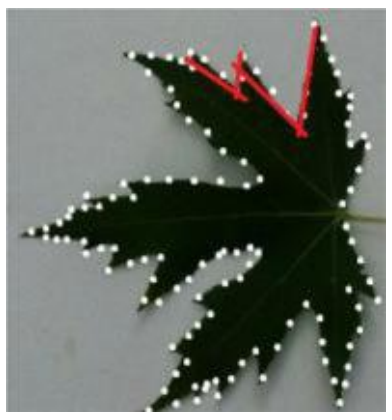 | Leaf shape | Smooth contour obtained with small number of vertices<br>It is varying in translation, rotation and scaling |
| Contour Characteristic points [76] | 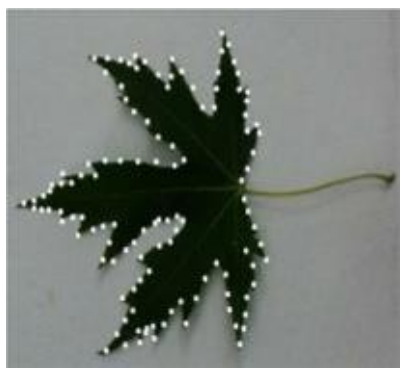 | Contour points selected depends on the curvature of contours | It is robust to translation, rotation and scale invariant |
| Active shape model [41, 77] | 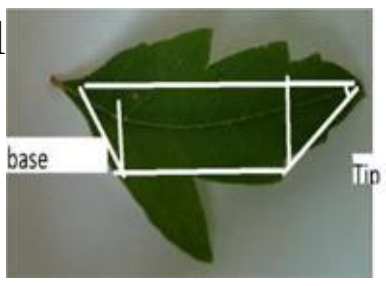 | Leaf tip | Finds leaf tip points and overlapping leaf tips<br>Used only for slender and thread-type leaves |
| Active polygonal model [42] | | Leaf tip and leaf base | Preserves leaf corners<br>Leaf tips vary in images, and there is damage to leaf corners |
| Contours of string [17, 43, 49, 79] | 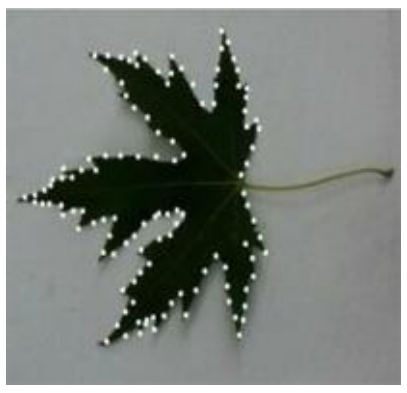 | Leaf margin | Semantically represents leaf margins<br>Leads to misclassification when there are imperfect leaf margins and overlapped leaves |
| Curvature scale space [16] | 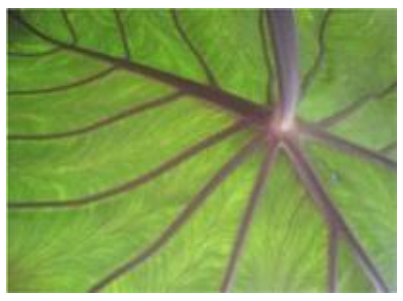 | Leaf venation | Finds the starting and ending points of leaves<br>Produces noise and is sensitive to rotation |
| Multiscale triangular area representation [23] | 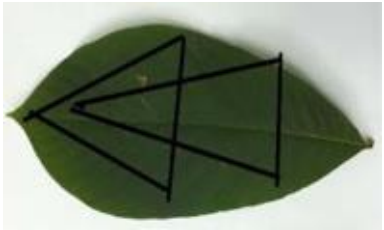 | Concavity and convexity of images | Affine invariant and robust to noise<br>Not scale invariant |
| Multi scale arch height descriptors [24] | 1-scale 2-scale and 4-scale<br>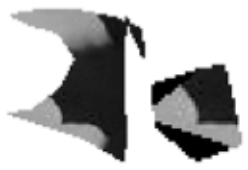 | Leaf margin | Measures the arch height of lobe-shaped and palmate-shaped leaves<br>Unsuitable for overlapped leaves<br>Normalization applied for scaling and rotation, taking up time |
| Multi scale bending energy [25] | 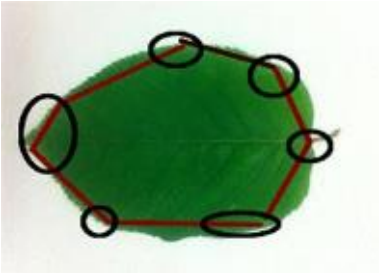 | Energy | Sensitive to local features of leaf shape contours<br>Provides low noise immunity |
| Curvelet transform [26] | 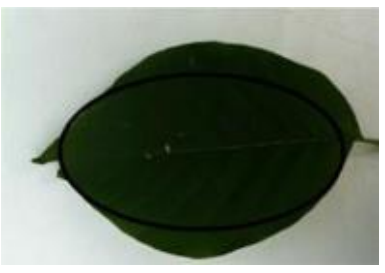 | Curvelet features | Useful for small objects<br>Unsuitable for natural images |
| Multiscale R-angle descriptor [27] | 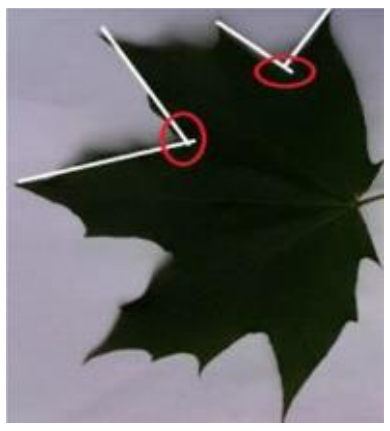 | Leaf Margin | Intrinsic to shape contour under translation, rotation and scaling<br>No need for normalization |
| Centroid contour distance, Angle code histogram [28–31] | 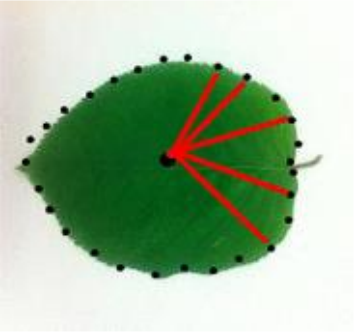 | Contour points and orientation angles | Invariant to translation and rotation<br>Used for compound, oblong and orbicular leaf shapes<br>Applicable only to leaf tip and base |
| Contour Key points [178] | 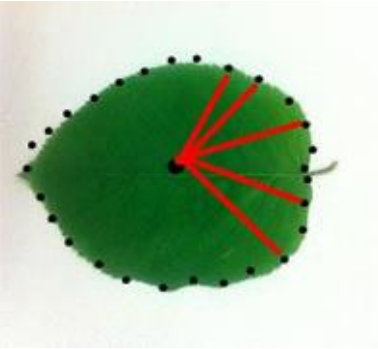 | Contour key points are extracted and represented as histogram bins by using fuzzy score | Solves intra class problem of same Species |
| Complex network Descriptor [78] | 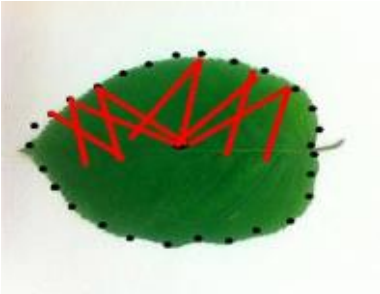 | Measures degree, joint degree of leaf boundary | Invariant to scaling and rotation<br>Noise tolerant |
| Geometric features [50–72, 138, 188] | 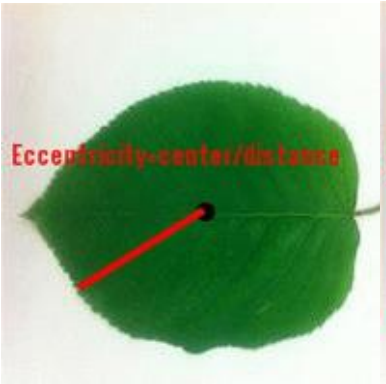  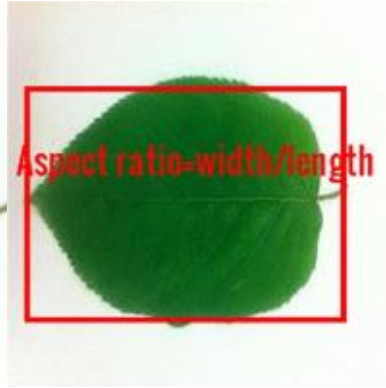  | Eccentricity<br>Aspect ratio<br>Leaf area<br>Leaf perimeter<br>Major and minor axis<br>Solidity | Low-dimensional features<br>Low computational costs<br>Minimal processing time<br>No accurate classification, because of similarities between interspecies and intraspecies |

Siricharoen et al. [100] used 13 textural features and 6 different Tamura's texture features for plant disease monitoring in a mobile cloud environment. The shortest path texture context [35] measures the shortest path along different orientations. Combining texture information and global shape information with local patches, the authors used gradient changes for lighting invariance. Wang et al. [101]. introduced a local n-ary pattern for texture classification which is rotation invariant and produces uniform patterns. However, it produces naturally high dimensional features. Wang et al. [102] used Local binary pattern in decomposed leaf images for extracting the characteristics of texture features of images on ICL and Swedish leaf databases. It is robust to noise, occlusion and clutter.

### 4.3 Textons

Textons are used to construct texton dictionaries created based on filter responses in spatial and frequency domains. For rotation invariant databases, the authors of [103] constructed a continuous maximum response descriptor to distinguish between and intra-class variations and a principal curvature descriptor for strong intra-class grouping ability. These techniques are useful for leaf databases with both interclass and intra-class variations. Minu and Thyagarajan [104] used texton with MPEG 7 visual features to recognize flower images. They also presented an ontology-based image retrieval system for asteroideae flower domain in their paper [105].

Guo et al. [106] classified rotation invariant texture by first finding out dominant orientation and then extracting anisotropic features by this orientation. They also proposed two statistical texton based methods to validate their approach. Anisotropic images change in appearance and rotate to produce good quality textures. The average and standard deviations of responses were computed in 8 different directions and a joint sort was used to find the local patch. These methods can be used to classify leaves in rotation invariant leaf databases. Table 5 shows some of feature descriptors used in leaf recognition.

## 5 Invariant Feature Detectors

Image transforms convert sets of images into a series of orthogonal images in the form of unitary matrices. The primary aim of transformation is to represent a unit image into a set of linear combination basic images, extract features like the edges and corners of images, and determine shift invariant rotations and scaling invariant images.

Pyramid Histogram-oriented Gradient (PHOG) [112] computes the local shape and global spatial information in leaf images. It extracts edge contour information and calculates histogram bins on each local bin. It operates on dense grid cells and is invariant to geometric and photometric variations, except object orientation. The power spectrum with harmonic analysis in [113] has applied TSO invariance, such as translation, rotation, scaling and mirroring based on Fourier descriptors. They introduced affine invariant harmonic analysis of radii spectrum for an affine invariant transform. It is calculated based on image moments. A redundant discrete wavelet transform [114] identifies orthogonal moments. Unlike other wavelet transforms, it does not consider all the input pixel values of images. It considers only odd pixels for scaling, including pixels for wavelet coefficients and reduces computational complexity. A polar Fourier transform (PFT) [115, 116] converts an original image into polar space so it is translation invariant, and as phase information is neglected, it is rotation invariant as well. The first magnitude value is normalized into scaling invariants, compared to other moment-based Zernike polynomials. They classified leaves using probabilistic neural network by incorporating shape, vein, colour and texture features with it and achieved 93.2% of classification accuracy compared to geometric and moment invariant features in their own databases. A log polar transform [117] was used with rotation and scale invariant features to classify different texture patterns. It follows point singularities and converts images into concentric circles. They stated that ridgelet transform was useful for texture classification and these features are rotational and scale invariants. They demonstrated that it provided 100% accuracy, an excellent result compared to the result produced by log polar transform on a rotational and scale invariant database of images. It is optimal to find only lines of the size of the image.

The Fourier–Mellin Transform [118] is a useful mathematical tool for image recognition as its resulting spectrum is invariant to rotation, translation and scale. The Fourier–Mellin descriptors are also invariant to the position of the object because they are derived from the energy centroid of an image, and it is transformed into a polar coordinate system that is invariant to the translation of the object. Squared moduli promote orientation invariance to phase the shift of the circular harmonics of images. The normalization permits both scale and intensity invariance. Thus, the Fourier–Mellin transform is invariant under translation, rotation, scaling and illumination changes.

Table 5 A summarization of texture, texton and LBP descriptors

| Feature extraction technique | Extracted feature | Advantages/disadvantages |
|---|---|---|
| Multi scale fractal dimension [81–83, 176] | Boundary and vein of leaf | Pros: discriminates between boundaries and patterns<br>Cons: Cannot discriminate between two objects with different patterns |
| Lie group of region structure [80] | Weed textures | Pros: measures self-similar structures<br>Cons: Small leaf dimensions |
| Lacunarity [84] | Spatial distribution of texture gap | Pros: identifies different image texture patterns<br>Cons: cannot measure invariant characteristics |
| Gabor filter [85, 86, 98] | Statistical features | Pros: extracts 13 different statistical measures |
| Boligon–Minkowski fractal dimension method [87, 88] | Texture | Pros: counts the number of boxes in spatial relationships<br>Cons: does not consider invariant features |
| Singular value decomposition [89] | Texture Characteristics | Pros: classifies texture characteristics on high-level factorization<br>Cons: provides good results in varying lighting conditions |
| Spatial dependency matrix (Gray Level Cooccurrence Matrix) [90, 107, 179] | Statistical features | Pros: measures skewness, entropy, standard deviation, and variance |
| PRICoLBP (Priority Co – occurrence Local Binary Pattern) [92] | Local curvature edge and contour<br>Information | Pros: applied to color images and is rotation invariant |
| LBP-HF (Local Binary Pattern Histogram Fourier) [93] | Uniform patterns using Fourier descriptors | Pros: stores all uniform patterns in 1 bin<br>Stores leaf interior and exterior information separately |
| MLBP (Modified Local Binary Patter) [94] | Statistical features | Pros: captures structural relationships between the gray values of the pixels in the neighbourhood |
| LBP with GLCM [95] | Multiscale texture features | Pros: no overlap between windows |
| Pi-LBP (Path Integral Local Binary Pattern) [96] | Texture | Pros: encodes cross-scale correlation |
| Pixel-based LBP [97] | Local and global information on texture | Pros: provides intra-class variability of pixels |
| Shortest-path texture context [35] | Texture and shape | Pros: invariant under lighting conditions |
| Local N-array pattern [101] | Texture | Pros: rotation invariant and produces uniform patterns |
| Continuous maximum response descriptor [103] | Textons | Pros: provides strong intra-class variability |
| Complex response filter [106] | Anisotropic features | Pros: produces good quality textures in complex responses |
| Transformation Spread function [108] | Shape | Pros: applicable for motion blurred image |
| Boosting Binary Key points [109] | Local patches | Pros: it requires less memory. More compact |
| Kernel Descriptors [110, 111] | Small patches | Pros: it improves patch level attributes instead of checking each pixel attributes |

Elliptic Fourier descriptors [119] were used for closed contours of leaflet edges to track the growth of velvet leaves. The misclassification that occurs in that method is due to the leaf plane orientation. The authors of paper [120] introduced a projection wavelet fractal descriptor which was used to reduce 2-dimensional features into 1-dimensional features, and to create sub-patterns of various features. The curves are non-self-correlated and circle projections were used there. It is rotation invariant and the projections are carried out with concentric circles. A minimum perimeter polygon [17] was applied to extract curvature descriptors outside the boundary, and the polygons were represented by a chain code. The algorithm produces inaccurate classification if there are too many straight lines, and an equal number of superfluous points along the boundary. Zernike moments [121] are used for the feature extraction of leaf shapes, given that leaves are irregularly shaped. It allows the extraction of shape vectors which are invariant to translation, rotation, scale, and skew and stretch options. Its higher-order polynomial produces global shape information, while the lower-order one offers local shape information. It improves accuracy overall.

Edge Angle (EAGLE) descriptor [122] was used to identify the angle relationship between lines of veins. The authors divided the entire image into 5 patches. The veins in each patch are modeled as lines. The method is limited to only 5 patches and performs no operations on a pixel-bypixel of an image. The Tchebichef moment invariants [123] were used to extract translation, rotation, and scaling invariant features. Legendre and Zernike's moments are orthogonal moments but the techniques produce a lot of information that is redundant. However, the Tchebichef produces less information redundancy and extracts information on moments in discrete orthogonal functions. It is used to extract pattern features from 2-dimensional images. Table 6 summarizes the invariant feature descriptors used in leaf recognition and, their advantages and disadvantages.

## 6 State-of-the-Art Classification Techniques

Plant species classification can be carried out by botanists easily, but computer-assisted systems cannot do so as easily. Consequently, plants are classified through leaf shape, vein, color and the texture of the leaves. Plant species are classified through different classifiers. A classifier requires two sets of data, a training set and a test set, but does not consider class relationships and the illumination invariance and positional invariance of images. Certain authors use manifold learning for classification since it preserves local neighbourhood structure, and high dimensional data is mapped into a low-dimensional structure. It also considers all illumination and positional challenges and processes noisy images. Compared to linear and supervised classifiers, manifold learning offers a good accuracy on plant species identification (Fig. 4). 6.1 Artificial Neural Networks

A neural network is a machine learning technique used for classification. The authors of the paper [124] identified disease in cotton, lemon and orange with the color feature and achieved 76.41% abnormality and 9.09% abnormality in leaf disease detection. Back propagation neural network (BPN) was used to classify half leaves based on the boundary tokens of shapes such as the angles and sinus of leaves in [125]. The authors examined 111 leaves of 14 different classes. It is a feed forward, self-adaptive network. Weights are adjusted based on the minimum mean square error. It takes longer time to train the network. Bagalkote et al. [126] used the BPN to classify grape varieties using texture and wavelet features and achieved 93.3% accuracy.

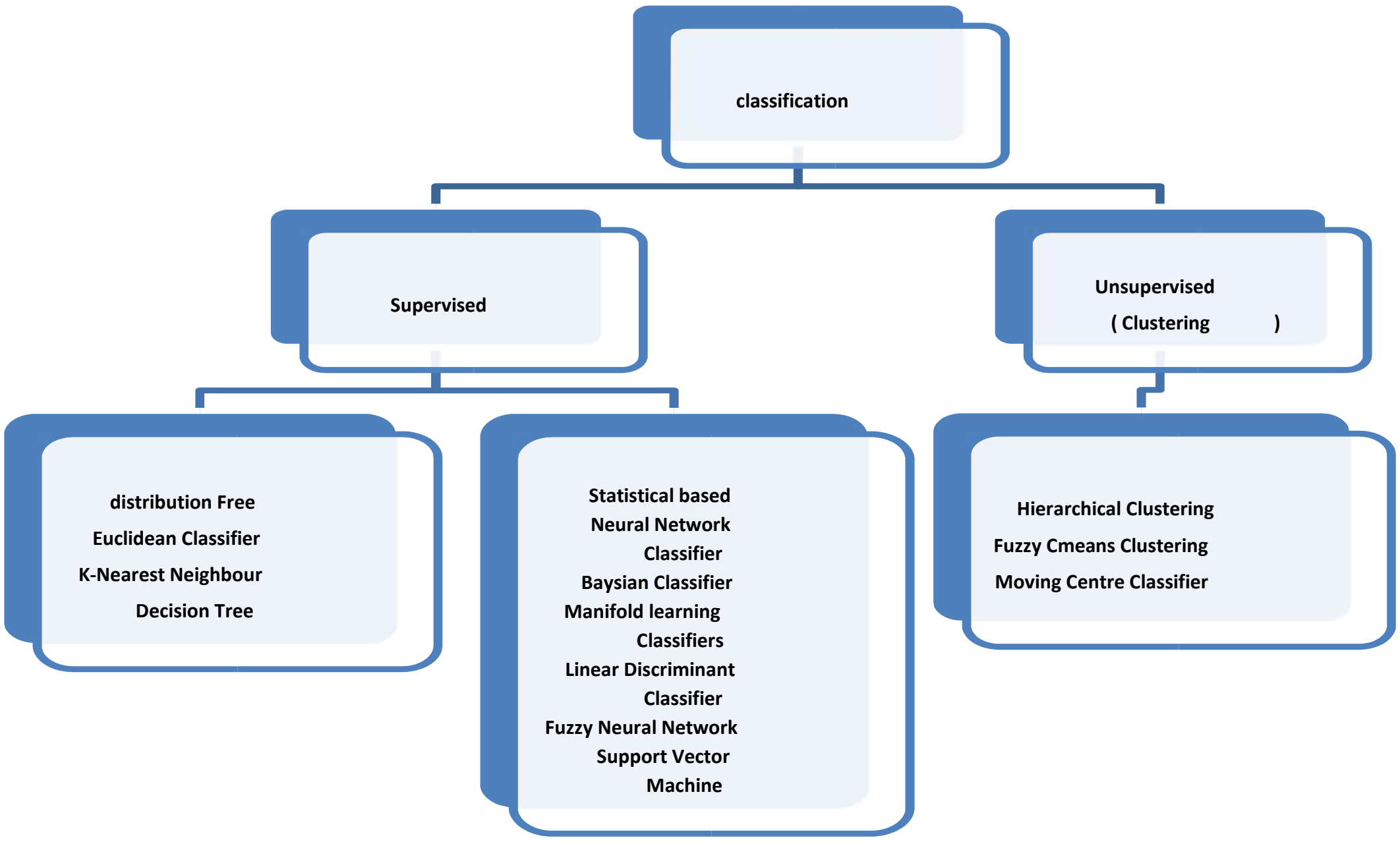

Fig. 4 General classification technique

Table 6 Summarization of invariant descriptors

| Feature extraction technique | Features | Advantages/disadvantages |
|---|---|---|
| PHOG (pyramid histogram-oriented gradient) [112, 186] | Edge contour information | Pros: invariant to geometric and photometric variations |
| Power spectrum with harmonic analysis [113] | TSO invariance | Pros: invariant to translation, rotation, scaling and mirroring |
| Redundant Discrete wavelet transform [114] | Wavelet features | Cons: considers only odd pixels for scaling |
| Polar Fourier transform [115, 116] | Phase information | Pros: translation and rotation invariant |
| Log polar transform [117] | Point singularities | Pros: rotation and scale invariant |
| Elliptic Fourier Descriptors [119] | Closed Contour of leaflet edges | Cons: misclassification occur s due to the leaf plane orientation. |
| Zernike moments [121] | Moment features on leaf shape | Pros: translation, Scale, rotation and skew invariant. |
| Tchebichef [123] | Moment Invariant | Pros: produces less information redundancy<br>Cons: extract pattern features from 2 dimensional images |

Anami et al. [127] used neural network to identify affected species of leaves based on color and texture features and identified 85% of affected vegetables and 80% of normal ones. Neural network was applied in [128] for plant disease classification and identification, based on the color co-occurrence texture features of the leaf. The BPN was used in [129] with the edge features for classification of leaves such as the neem, pine and oak and achieved 90.45% classification accuracy. The authors of the paper [130] used the BPN to classify the night jasmine, arka (blue madar), mango, neem, and shigru (moringa/drumstick) and achieved 85% accuracy.

The radial basis function is a three-layer feed forward network used for image classification, and produces faster training speeds compared to the Multilayer Perceptron (MLP). Sumathi et al. [98] used this approach to classify 90 samples and achieved 85.93% accuracy with a minimum mean square error. This method works well on spherical and regularized linear spaces. Akif and Khan [65] used the ANN to classify 817 samples of 14 different trees with morphological features, utilizing the Fourier descriptor and shape-defining features and achieved 96% accuracy. The author of the paper [60] used the ANN to classify 80 leaf images with 10 different classes with 7 different morphological features and achieved 98.8% classification accuracy. The researchers in the paper [89] used BPN to train herbs of 600 training samples and 1400 test samples with the texture feature and achieved an average accuracy of 98.9%. The authors of paper [131] accounted the single hidden layer feed forward network for classification. There is no need to use a kernel function to approximate the weights, given that it updates the weights randomly for fixed bias inputs. It has no control parameters such as learning rates, learning epochs and stopping criteria. They achieved 98.17% classification accuracy.

Chaki et al. [26] designed a neuro-fuzzy system with a back propagation multi-layered feed forward network to classify 930 images of 31 classes and achieved 97.6% accuracy. Because neuro-fuzzy system uses the probability of classes, to avoid problems in the ANN, fuzzy C-means clustering works by assigning each membership to each data point corresponding to other data points which belong to more than one cluster and it gives good results for overlapped datasets. In k-means clustering, data points belong to more than one cluster center, but here they are assigned. The authors of papers [42, 43] used this algorithm to classify species of plant databases with the specified margin structure. Balasubramanian et al. [86] formulated the fuzzy relevance vector machine to classify 60 categories of leaf images with shape and texture features. This method helps to select the optimum features of an image, achieving 99.87% accuracy. Sharma and Gupta [132] developed a system to classify agriculture and Ayurvedic plants using a multilayer feed-forward network with back propagation algorithm. They tested their system with 440 leaves of 16 classes and obtained classification accuracy greater than 90%.

### 6.2 K-Nearest Neighbours

The K-Nearest Neighbor (K-NN) is a simple technique used to classify objects with the closest training samples in feature spaces. Images are classified, based on the majority voting of its neighbors. Du et al. [120] used the KNN for classifying plants of 30 species with 2422 image samples based on edge, vein and ring projection fractal wavelet features as a new shape feature, and achieved 87.14% of classification accuracy with the size of the feature vector as 20. The authors of the paper [134] used the KNN to classify 300 images of 10 classes of leaves using 13 geometrical features and achieved 80% accuracy. The researcher in [135] used the KNN classifier to classify 100 leaf species with 1600 samples using the leaf margin as a feature and an interior texture histogram of 64 different feature vector images, and achieved 75.5% of classification accuracy.

Jose et al. [40] used the KNN for Costa Rican plant identification in the Flavia dataset using the features of the 0.5 HoCS (Histogram curvature Scale Space), LBPV (Local Binary pattern Variance), R1P8 (1 rotation with 8 pair of neighbourhood pixels), and R3P16 (3 rotations with 16 pairs of neighbourhood pixels) with $k = 10$ and achieved an accuracy of 99.1%. The authors in [120] used KNN with fractal dimension of the RPWFF to classify a total of 2422 images of 30 different species and achieved 87.14% of classification accuracy. Zhao et al. [37] used KNN to classify the Swedish, ICL, Smithsonian and Plummers Island datasets with a pattern counting approach and achieved 97.07, 73.08, and 72.28% classification accuracy respectively. The researchers of the paper [94] used the KNN with texture features of leaf for classifying intracluster variations of the Flavia leaf dataset and achieved 97.55% accuracy. Arunpriya et al. [133] experimented with fuzzy inference system, radial basis function network and K-nearest neighbour classifiers and classified tea species using leaf images and came to a conclusion that fuzzy inference system obtained better accuracy and took less time for execution compared to other two classifications. Elhariri et al. [136] applied LDA and RF to classify tree species of the UCI Machine Learning Repository using the texture, shape, and vein features.

### 6.3 Moving Center Hyper spheres

A Moving Center Hyper sphere classifier (MCH) [137] was proposed for high-dimensional features. In the KNN and neural network, the classification of plants is a laborious and space-consuming process. In the MCH, however, the features are arranged as n-hyper spheres. Using this classifier, 1200 leaf samples of 20 classes were tested with 23 moment invariant features and achieved 92.6% accuracy.

### 6.4 Bayesian Classifiers

The Bayesian classifier, a simple probabilistic classifier based on Bayes' theorem, computes the posterior probability for the targeted output. The researchers in paper [116] used the Bayesian classifier with the Fourier descriptor feature to classify 100 different kinds of leaves and achieved 88% accuracy. They used the linear classifier with the features of the polar Fourier transform, color, vein and 20 features of lancularity, solidity and convexity of shape for classifying the Flavia and Foliage databases and achieved 95.94 and 93.25% accuracy respectively.

### 6.5 Support Vector Machine

The support vector machine (SVM) is a linear classifier. The process of classifying leaf species calls for a multiclass classifier, because multiple leaf species are identified by multiclass SVMs. Compared to the neural network classifier, it performs better because of its selection of kernels. No prior training is called for, though it involves a huge number of images. In [38] the authors used SVM-RBF (Radial Basis Function) kernel to classify leaves of the Leaf snap database. The RBF kernels automatically produce a number of support vectors, centers, and weights during the training. The authors of the paper [114] used the SVM with wavelet features of images to classify ornamental plants with 95.83% accuracy. SVM classifier was used to classify species of Annona squamosa and Psidium guajava in [139] with Hu moments, achieving an average accuracy of 86.66%.

A multiclass SVM [93] was applied on the Australian Federal dataset, Flavia, Foliage, Swedish and Middle European datasets with the texture features and Fourier transform descriptors and combined the features of interior and boundary descriptors extracted, and achieved classification accuracy of 100% in the AFF, 99.7% in Flavia, 99.8% in Foliage, and 99.2% in MEW (Middle European Woody) datasets. The authors of papers [111] used one versus all SVMs in the Flavia dataset with kernel level descriptors [110, 111] and achieved an average accuracy of 97.5% (1585 training images of 32 species and 320 testing images), and 58% with Image CLEF 2013 (7525 training images of 70 species with 1250 testing images. The authors of the paper [139] proposed relative sub-image features to be extracted from the whole image. They extracted 300 features from each image and used a support vector machine as classifier and achieved 97.25% accuracy.

SVM classifier with the fractal dimension of the leaf shape with its lancularity features was used in [84] to classify 626 images of the Flavia dataset with an average accuracy of 95.048%. The SVM used to classify the Flavia and Swedish datasets with the features of the HOG and Zernike moments [140] with 40 samples provided an average accuracy of 97.18 and 98.13% respectively. The SVM used with a combination of fractal descriptors to identify the Brachiaria species [87] achieved a classification accuracy of 93%. The SVM was used to classify the ICL (Intelligent Computing Laboratory) dataset with interclass similarity and achieved 90% accuracy when the features of PCNN entropy, Hu and moments were used [73]. They used the ICM [75] and center distance sequence with the SVM RBF kernel on the Flavia dataset to achieve 97.82% accuracy. Narayan and Subbarayan [149] extracted color and boundary sequences, and optimized the feature extraction using genetic algorithm (GA) to improve the performance based on matching accuracy. They used SVM for classification.

### 6.6 Principal Component Analysis

The PCA is an algebraic technique used to select important correlated variables from images. Glozarian and Frick [54] used the PCA to classify species of different grasses such as wheat, rye and brome grass by extracting the shape, color and texture features of images and achieved 88 and 85% for Wheat and brome classification accuracy. The PCA with textural features extracted by the gray-level cooccurrence method was used to classify 390 leaves with 13 different kinds of plants and achieved 98% accuracy in [107]. Mebatsion et al. [141] used PCA for classifying different grain types with achieved accuracy of 99%. The authors also extracted shape, texture, and color features from leaf images in [150] and optimized i.e. selected a subset of features using genetic algorithm and Kernel based Principal Component Analysis (KPCA) to improve the accuracy of classification.

### 6.7 Random Forest

An ensemble classifier, the random forest is used to construct a large set of trees at random. It runs efficiently on large databases, handles a large number of input variables without variable deletion, effectively estimates missing data, and maintains the accuracy of a classifier. It gives proximities between pairs of classes and, further, estimates crucial features automatically. During multiclass classification, if some data are missed, it leads to an imbalance in the data concerned. To resolve this problem, a direct ensemble classifier [142, 143] is used for an imbalanced multiclass learning classifier. It is a combination of the 1-nearest neighbor and Naive Bayes or the K-nearest neighbor and Naive Bayes classifiers.

### 6.8 Convolutional Neural Networks

All classifiers handle a small number of images, except the CNN (convolutional neural network), which handles large set of images. For all classifiers, feature extraction is a separate space, since they cannot directly extract features from images. The CNN, however, extracts features directly from the images in question, disregarding illumination, lighting, shadowing or skewness. It is not rotation invariant but translation invariant, needing similar-sized images for classification. The CNN was used to classify large sets of images by Dyrmann et al. [144] and they trained 10,413 images of 22 species, achieving a classification accuracy of 86.2%. The authors of the paper [145] used the CNN to identify 13 different plant diseases and achieved 96.3% accuracy. The researchers of the paper [18] applied the CNN to identify legume species of soya bean, white bean and red bean using the vein morphological features of 422 images of soybeans, 272 red beans and 172 white beans leaves and achieved an average recognition accuracy of 96.9%.

### 6.9 Manifold Learning

Plant leaf data are nonlinear and high-dimensional in nature. Manifold learning is a kind of nonlinear dimensionality reduction technique which discovers the nonlinear structure of the data and can be applied to leaf classification. Hu et al. [146] used multi-scale distance matrix (MDM) to extract global information from an image. It is invariant to translation, rotation, scaling and bilateral symmetry. But it is a time-consuming process, producing point wise matching of an image. They got 91.3% classification accuracy in the ICL and Swedish Leaf Dataset using the Linear Discriminant Classifier.

A supervised locality projection analysis (SLPA) was proposed to classify the ICL and Swedish leaf datasets in [147]. The SLPA was used to project a high-dimensional feature space into a two-dimensional feature space in which the projection is carried out by intra-class and interclass separability using labeled samples. They tested it with 50 samples of data of 11 species and achieved 96.33% accuracy. The authors in the papers [148] proposed orthogonal locally-discriminate embedding to consider the intrinsic manifold structure of leaves and they reviewed the local neighborhood preserving the structures of leaf images. They achieved 91.6% classification accuracy when used with 480 leaf images. Cem Kalyoncu et al. [64] used a local discriminant classifier to classify the Flavia and Leafsnap datasets. Since the LDC contains weights for the feature vector and neglects certain irrelevant features they achieved 99.1% classification accuracy. Table 7 summarizes the classification techniques used to classify the leaves of various plant species.

## 7 Comparison Analysis of Leaf Datasets

All the datasets discussed in this study fall into three categories: scan, pseudo-scan and photos. Herbarium datasets are obtained by using images with a simple, plain background. Some images show stalks, while others show blades without petioles. Here we discuss certain publicly available leaf datasets.

### 7.1 The Swedish Leaf Dataset

Soderkvist et al. [158] created the Swedish Leaf Dataset [159] for Linkoping University and the Swedish Museum of Natural History. The dataset contains scanned images of 15 tree species with 75 leaves per species, against a plain background. It is a challenging dataset because of its interspecies similarity. The leaves in the dataset are arranged manually. Only a single side of a leaf with petioles is captured. The petioles have lots of discriminant information. During processing, however, the petiole is removed. The leaves are all of good quality and with no holes. Almost all the authors [92, 93, 113, 140, 160] who worked on the Swedish Leaf Dataset applied the support vector machine to classify the species. The SVM, a supervised learning model for classifying labeled images, is used because the Swedish Leaf Dataset has already labeled 15 tree species (Fig. 5).

The authors used shape and texture to achieve the highest accuracy ratio of 99.38% [92]. All the images are scanner-based moment invariants, scanned with a plain background. The authors of [113] used the KSVM classifier with the features of translation, scaling, rotation and mirroring invariant descriptors with the KSVM, given that kernels are useful for handling high-dimensional feature vectors, particularly when the feature vectors are not linearly separable. They used TSO/A (translation scaling, affine invariant) features based on the HATSIS (Harmonic analysis of texture invariant); consequently, the feature vector cannot be divided into linearly separable patterns. So we cannot divide the feature vector into linearly separable. They used the KSVM and achieved an accuracy of 97.87%.

The next most widely used classifier, following the KSVM, is the KNN nonparametric classifier that classifies unknown samples based on their K-nearest neighbor among the training samples. The difficulty with the KNN is choosing a K-value, but in [24] Multiscale arch height descriptor was used and achieved an accuracy rate of 96.21%, though the authors achieved 99.25% accuracy using the fuzzy KNN classifier in [102] that assigns membership as a function of the object distance as a classifier. It is computationally faster when compared to the KNN. The researchers in [122] used EAGLE (Edge Angle) descriptor with the bag-of-visual-words model and achieved an accuracy rate of 94.9%. Xiao et al. [197] used histogram of Oriented gradient feature with maximum margin criteria for classifying Swedish leaves and he achieved higher recognition rate even though the species contains same blades.

### 7.2 The Flavia Dataset

This dataset contains 1907 leaf images of 32 different species and 50–77 images per species. The leaves were collected from 32 common plants in the Yangtze Delta (where Shanghai is) of China. The dataset provides highlyconstrained leaf images against a white background where no stem is present. This dataset only covers 32 species with a single training image, and can be downloaded from [161] (Fig. 6).

The most widely used classifiers include the SVM, KNN, BPN, Navie Bayes and LDA classifiers. The SVM used with the RBF nonlinear kernel [75]. They extracted entropy and the contour distance sequence using an intersecting cortical model. Here, the number of observations is more, when compared to the number of features. Consequently, the authors achieved the highest recognition rate of 97.82%, compared to [140]. Sule et al. [93] claim in their report that they achieved the highest recognition rate of 99%, given that all species in the Flavia dataset exhibit high interspecies similarity, facilitating the extraction of texture features for classifying the species, thus culminating in a recognition rate of 99%.

Since morphological and geometrical features are not suited to the Flavia dataset because of its interspecies similarity, [24, 56, 64, 162] achieved less than 90% accuracy. However, in [67] geometrical features incorporating a neuro fuzzy classifier with the added advantage of its humanlike reasoning style and linguistic model, achieved an accuracy of 97.5%.

Table 7 Summarization of classification technique

| Classifier | Feature | Dataset and accuracy |
|---|---|---|
| Artificial neural network | Colour | Cotton, lemon, orange disease<br>76.41% [124] |
| | Fourier Descriptors shape defining feature<br>Morphological features | 95% [65] |
| | Morphological feature | 98.8% [60] |
| | Texture | Herbs<br>98.9% [89] |
| | Hu, shape, texture | 100% [151] |
| Back propagation neural network | Shapes, Angles and sinus of leaves | 111 leaves with 14 species [125] |
| | Texture and wavelet feature | Grape varieties 93.3% [126] |
| | Colour | Affected and Unaffected Vegetables [127] |
| | Texture of Colour Co-occurrence | Disease Identification [128] |
| | Edge Features of leaf | Neem, pine oak 90.33% [129] |
| | Leaf Margin | Jasmine, arka, mango, neem and shigru 85% [130] |
| | Morphological features | 450 leaves of 16 classes in Ayurveda and agriculture 90% [62] |
| | Texture | Foxtail, crabgrass, velvet leaf, morning glory 97% [177] |
| Fuzzy based classifiers | Statistical features of leaf | 97.6% [26] |
| | Texture, shape, colour | 99.87% [86] |
| K-nearest neighbour | Edge, vein, ring projection wavelet feature | 87.14% [120] |
| | Geometrical features | 80% [134] |
| | Leaf Margin? texture | 75.5% [135] |
| | Texton | Costa Rican Flavia Dataset 99.1% [40] |
| | Texton | 87.14% [120] |
| | Texture | ICL-97.07% Plumber-72.8% Simthsonain-73.08% [37] |
| | HoCS, contour, colour, curvature | Flavia 99.61% [39] |
| | Texture | 97.55% [94] |
| | Run length sequence | 93.17% [152] |
| | Contour—amplitude frequency descriptor | Swedish-89.6% ICL-91.6% [198] |
| Moving centre classifier | Moment invariant | 92.6% [137] |
| Bayesian classifier | Fourier descriptor | 88% [116] |
| Support vector machine | HoCS | Leafsnap [38] |
| | Wavelet features | Ornamental Plants 95.83% [114] |
| | Fourier and texture | Australian Federal dataset-100%, Flavia-99.7%, Foliage-99.8%, Swedish and Middle European datasets-99.2% [93] |
| | Kernel level descriptor | Flavia-97.5% [110, 111] |
| | Hu moments | Annona Squamosa and Psidiuguajava, 86.6% [139] |
| | Lanculariity | Flavia-95.048% [84] |
| | HOG? Zernike moments | Sweedish-97.18%<br>Flavia-98.23% [140] |

| Classifier | Feature | Dataset and accuracy |
|---|---|---|
| | Fractal descriptors | Bracharria-93% [87] |
| | PCNN entropy, Hu moment | Flavia-90% [73] |
| | ICM-center distance | Flavia-97.82% [75] |
| | TSO invariant | Sweedish-92.27% [113] |
| | Shape, texture, morphology, colour | 2050 leaf images from Flavia and Ficus deltoidea, Citrus, and Acanthaceae plants 95.53% [181] |
| | Geometrical features—aspect ratio, slice ratio, radius ratio, ellipse equilibrium, circle equilibrium | 500 leaf images-95% [195] |
| | Shape, color, texture | ImageCLEF12-126 tree species 81% [199] |
| Support vector machine (one vs all) | Geometric + vein + Fourier descriptor | Flavia-87.4 [153] |
| Principal component analysis | Shape, texture, color | Wheet, rye, broom grass-88 and 85% [54] |
| | Texture | 98% [107] |
| Convolutional neural network | Texton | 10,413 images of 22 species-86.2% [144] |
| | Color | 13 different kinds of plant species diseases-96.2% [145] |
| | Vein and morphological features | Soya bean, red bean, white bean-96.9% [18] |
| Supervised locality projection | Shape, texture | ICL<br>Sweedish [147] |
| Orthogonally local discriminant embedding | Texture | 91.16% [148] |
| Dynamic time warping | Leaf margin | 90% with 100 different species [154] |
| Dictionary based Learning Model with sparse representation of Bag of visual words | Texture | Flavia-95.47% [155] |
| | SURF | Flavia-95.94% [156] |
| Fisher vector [157] | Local patches | – |
| Deep belief network (DBN) | Shape, texture, Hu moments | Flavia-220 species-93.9% [189] |
| Generalized recognition neural network | Geometric and morphological features | 10 species-100% [190] |
| Euclidean minimum distance classifier | Zernike moments | 50 types of leaves-84.66% |
| | Histogram of oriented gradients | 50 types of leaves-92.67% [191] |
| Learning vector quantization +radial basis function | Texture | 98.7% [192] |
| Map reduce algorithm | Texture | Hierarchical big database-91% [196] |

## 7.3 The ICL Dataset

The ICL dataset contains isolated leaf images of 220 plant species, with individual images ranging from 26 to 1078 per species. The leaves were collected from the Hefei Botanical Garden in Hefei, the capital of the Chinese Anhui Province [163] (Fig. 7).

The authors of [137] used the SVM classifier with the texton features and achieved higher classification accuracy of 97.73%, compared to the classification accuracy of 95.87% achieved in [75] with entropy feature. When the leaf margin was used as a feature, with orthogonally locally linear embedding of a manifold learning classifier, the authors of the paper [164] achieved an accuracy of 94.11%. Wang et al. [147] used labelled features and supervised locality projection to achieve a classification accuracy of 97.54%. Wang et al. [102] achieved the highest recognition rate of 98.03% compared to all the K-nearest neighbour classifiers because they used fuzzy K-nearest neighbour which acts like a human decision-making system because of its linguistic model.

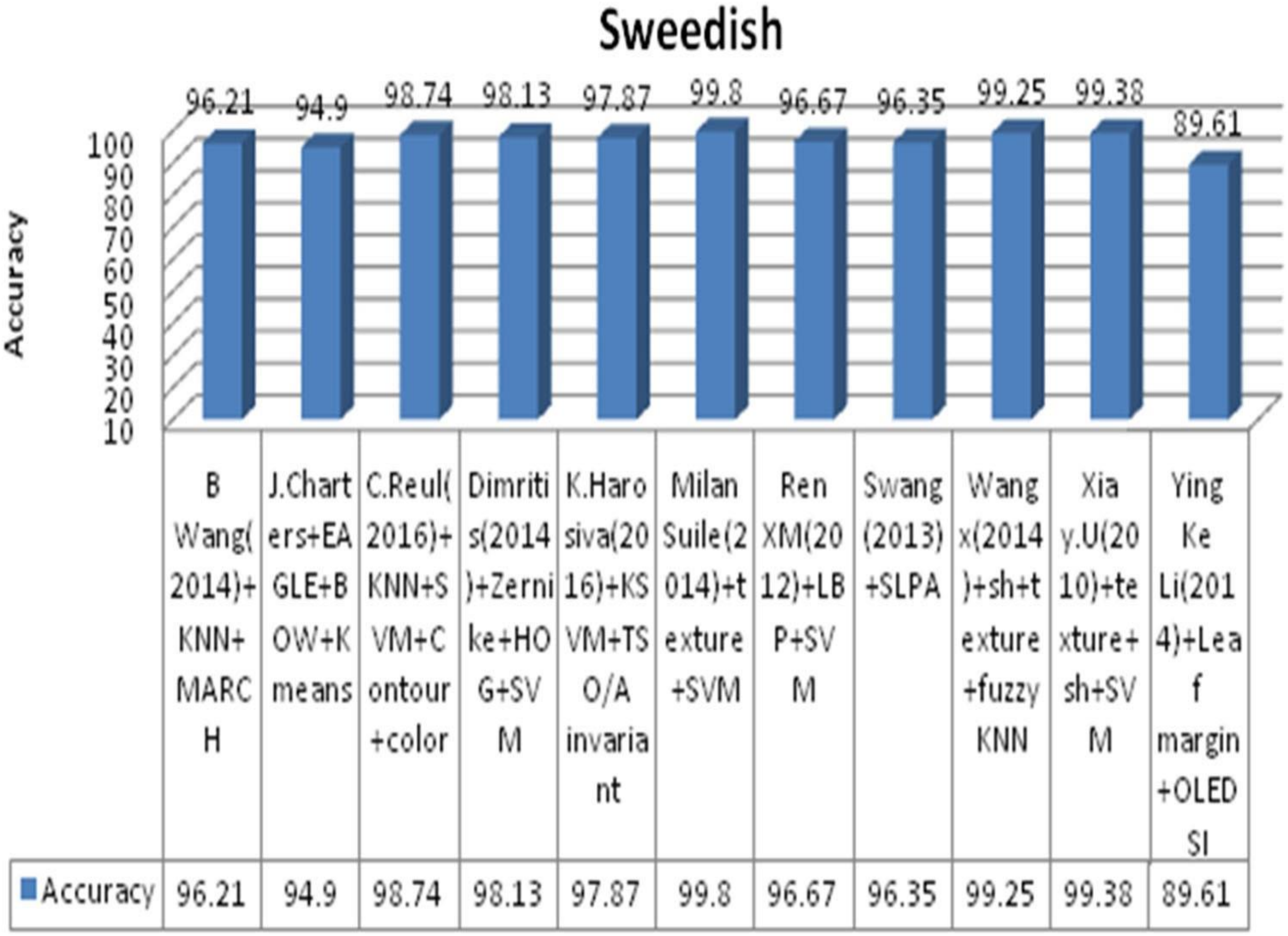

Fig. 5 Comparison analysis of different classifiers on Swedish database

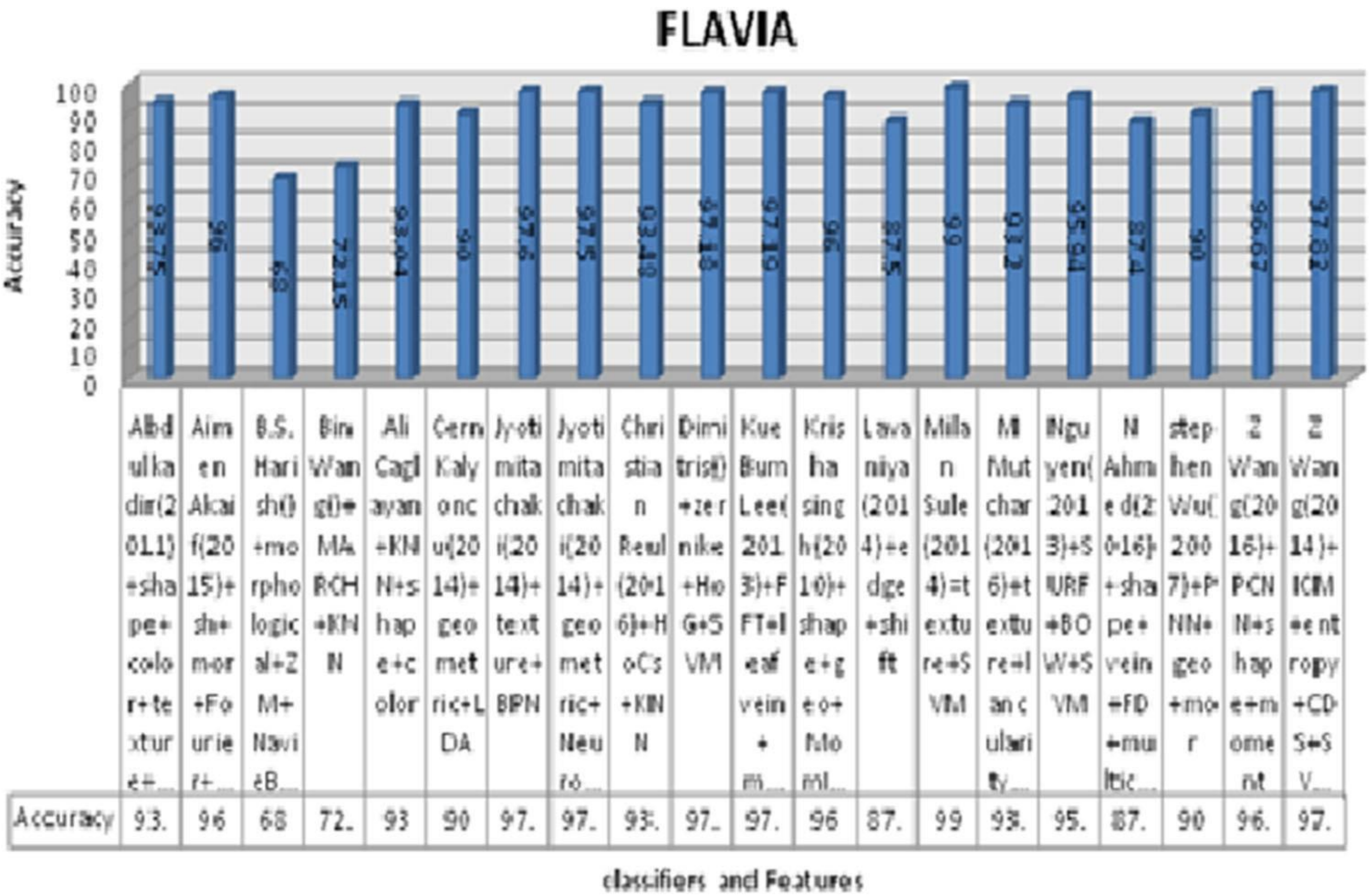

Fig. 6 Comparison analysis of different classifiers on Flavia database

## 7.4 The UCI Machine Repository

This dataset contains 340 species, each with 10 leaves. The dataset was created by Pedro et al. [166] using leaf specimens collected by Rubim Almeida da Silva at the Faculty of Science, University of Porto, Portugal. The dataset can be downloaded from [165]. Silva et al. [166] extracted eccentricity, aspect ratio, elongation, solidity, stochastic convexity, isoperimetric factor, maximal indentation depth, lobedness, average intensity, average contrast, smoothness, third moment, uniformity and entropy features with Linear Discriminant Analysis classifier and achieved a classification accuracy of 87%. Since these features are highly correlated to the species, we are consequently required to select appropriate features for the UCI Machine Repository.

## 7.5 The Austrian Federal Forest (AFF) Dataset

The AFF contains 134 leaf photos of Austrian 5 broad trees with the plain background. The Austrian Federal Forests (AFF) datasets comprise images of trees, leaves, bark and needles [167]. Fast Scale rotation invariant and texture features were used in [93] to perform a 10-fold cross validation with which a recognition accuracy of 97.32% was achieved.

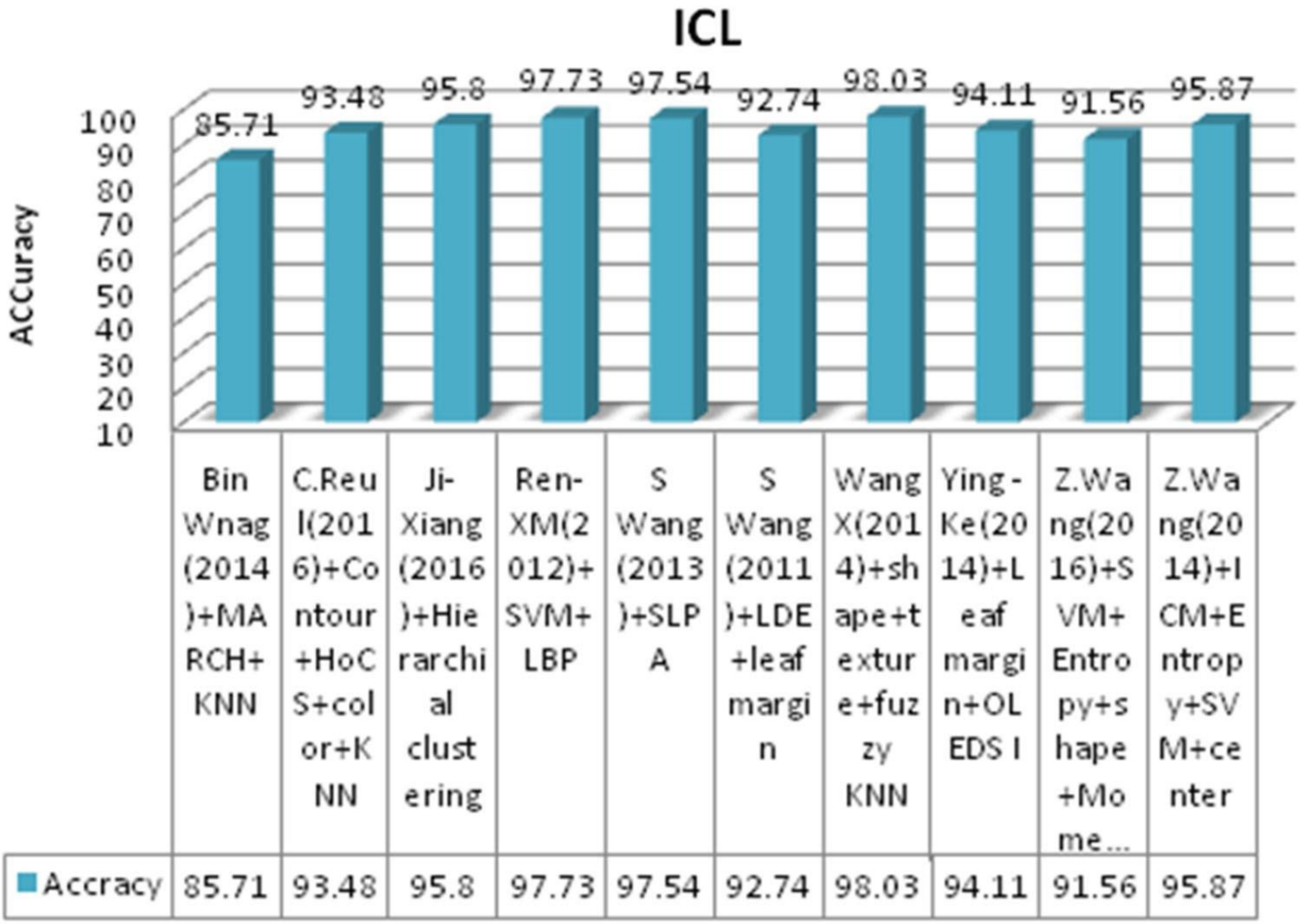

Fig. 7 Comparison analysis of different Classifiers on ICL database

## 7.6 The Smithsonian Leaf Dataset

The Smithsonian Isolated Leaf Database contains 343 leaves from 93 species. The Smithsonian has built a digital collection of specimens and provided the means to access it with text and photos of plants. The researchers created a system to extract leaf models with unknown samples. The images in question are taken from an indistinct lighting background and are not flattened well. It can be downloaded from [168]. Ling et al. [35] extracted the shape context from leaves and measured the distance between two species for classification.

## 7.7 Leaf Snap Dataset

This database can be downloaded from [169] Lab images, consisting of high-quality images taken of pressed leaves from the Smithsonian collection. These images appear in controlled back-lit and front-lit versions, with several samples per species. The 7719 Field images consist of ‘‘typical’’ images taken by mobile devices (mostly iPhones) in outdoor environments. These images contain varying amounts of blur, noise, illumination patterns, and shadows. The dataset currently covers all 185 tree species from the Northeastern United States. Kumar et al. [38] developed a mobile electronic field guide for the Leafsnap database. They extracted the features of the curvature histogram and IDSC descriptors with the KNN classifier and achieved 96.8% classification accuracy. Kalyoncu et al. [64] used geometric features with the Linear Discriminant Classifier to achieve 90% accuracy.

## 7.8 The Middle European Wood Database

The MEW (Middle European Woods) database, originally named LEAF, was created for experiments to do with the recognition of woods by shape of their leaves. It contains leaves of wood species growing in the Czech Republic, both trees and bushes; native, invasive and imported (only those imported species common in parks are included). The leaves were scanned with a 300 dpi, thresholded (binarized), pre-processed (denoised and cleaned) and saved in PNG format. The name of each file includes the Latin name of the species and the label of the sample. The database can be downloaded from [170]. Novotny et al. [171] used 151 tree species with at least 50 leaves per species. They tested compound leaves and revealed the differences between branches with compound leaves and those with pinnately compound leaves. They extracted features such as image moments, Fourier descriptors and leaf size and classified using the KNN classifier and achieved 88.9% accuracy. The authors of the papers [39] used the Histogram Over Curvature scale feature with the 1 Nearest neighbour classifier, obtaining an accuracy of 95.66%. Cerutti et al. [184] used 1000 compound leaf images of 17 European tree species.

## 7.9 Pl@ntNet

The database covers a large number of wild plant species collected from Western Europe and North America and contains 10,000 plant species. It can be down loaded from [172]. Aptoula et al. [66] used Image CLEF2012 [182, 183] leaf dataset with 10,000 leaf samples of scan and scan-like images. Extracting features of circular covariance and a morphological histogram from 6270 samples of 91 species, they achieved a classification accuracy of 56.09%.

Yahiaouri et al. [173] used the leaf margin as a descriptor with the K-Nearest Neighbor classifier and achieved 77% accuracy. Ceruttiet al. [43] used the leaf margin as a descriptor. They tested 5668 leaf images of 80 tree species with fuzzy C-means clustering and achieved accuracy of 88%. Liu et al. [174] used the Random Forest classifier with the shape feature and achieved an accuracy of 65%. Zhao et al. [112] used the ImageCLEF 2012 Leaf database of 126 tree species with the hog, color, and texture feature of 4870 scan-like and scan photos of 2500 images for training. Joly et al. [175] developed a plant identification system based on social image data of France and it covers 2200 species of leaves. Mouine et al. [185] developed a plant identification system with the features of the leaf margin and salient points of leaf. He used 1819 scan like images for training and 907 images for testing.

## 8 Conclusions

In this paper, we have discussed a number of leaves species identification methods for a domain knowledge-based system. It should be noted that no single method is adequate enough to identify a species effectively. Depending upon the problem stated, an appropriate feature extraction method is to be selected, given the plants vary with geological habitats and can be affected by photometric and geometric conditions. During storage, plant images may suffer from noise, moments and the size of the images themselves. We have here discussed a variety of methods to extract the different features of plants, including shapes, colors, textures, moments, as well as geometric and photometric invariants, as features are fundamental in computer vision to classify images of any species. Similarly, we have discussed classification techniques such as machine learning, manifold classifiers, ensemble classifiers and linear classifiers. Based on the discussion and need, we select an appropriate feature, feature extraction method and a classifier—efficient in terms of both the space and time complexity of images—to help in the identification of large plant species.

## Conflict of interest: The authors declare that they have no conflict of interest.